\documentclass[runningheads]{llncs}

\usepackage[T1]{fontenc}
\usepackage{graphicx,verbatim}
\usepackage{amsmath}
\usepackage{multirow}
\usepackage{pifont}
\usepackage{amssymb} 
\begin{document}
%

\title{SIMPLER: H\&E-Informed Representation Learning for Structured Illumination Microscopy}

\titlerunning{SIMPLER}
%
\author{%
Abu Zahid Bin Aziz\inst{1,2}\thanks{Corresponding author: zahid.aziz@sci.utah.edu} \and
Syed Fahim Ahmed\inst{1,2} \and
Gnanesh Rasineni\inst{1,2} \and
Mei Wang\inst{3} \and
Olcaytu Hatipoglu\inst{4} \and
Marisa Ricci\inst{4} \and
Malaiyah Shaw\inst{4} \and
Guang Li\inst{4} \and
J. Quincy Brown\inst{4} \and
Valerio Pascucci\inst{1,2} \and
Shireen Elhabian\inst{1,2}
}

\authorrunning{Aziz et al.}

\institute{%
Kahlert School of Computing, University of Utah, Salt Lake City, UT 84112, USA
\and
Scientific Computing \& Imaging Institute, University of Utah, Salt Lake City, UT 84112, USA\\
\email{zahid.aziz@sci.utah.edu, u1419916@utah.edu, gnanesh.rasineni@utah.edu, valerio.pascucci@utah.edu, shireen@sci.utah.edu}
\and
Instapath Inc., Houston, TX 77021\\
\email{mwang@instapathbio.com}
\and
Tulane University, Department of Biomedical Engineering, New Orleans, Louisiana, United States\\
\email{ohatipoglu@tulane.edu, mricci1@tulane.edu, mshaw5@tulane.edu, gli3@tulane.edu, jqbrown@tulane.edu}
}


  
\maketitle              
%
\begin{abstract}

Structured Illumination Microscopy (SIM) enables rapid, high-contrast optical sectioning of fresh tissue without staining or physical sectioning, making it promising for intraoperative and point-of-care diagnostics. 
Recent foundation and large-scale self-supervised models in digital pathology have demonstrated strong performance on section-based modalities such as Hematoxylin and Eosin (H\&E) and immunohistochemistry (IHC). However, these approaches are predominantly trained on thin tissue sections and do not explicitly address thick-tissue fluorescence modalities such as SIM. When transferred directly to SIM, performance is constrained by substantial modality shift, and naive fine-tuning often overfits to modality-specific appearance rather than underlying histological structure.
We introduce SIMPLER (Structured Illumination Microscopy-Powered Learning for Embedding Representations), a cross-modality self-supervised pretraining framework that leverages H\&E as a semantic anchor to learn reusable SIM representations. H\&E encodes rich cellular and glandular structure aligned with established clinical annotations, while SIM provides rapid, nondestructive imaging of fresh tissue. During pretraining, SIM and H\&E are progressively aligned through adversarial, contrastive, and reconstruction-based objectives, encouraging SIM embeddings to internalize histological structure from H\&E without collapsing modality-specific characteristics.
A single pretrained SIMPLER encoder transfers across multiple downstream tasks, including multiple instance learning and morphological clustering, consistently outperforming SIM models trained from scratch or H\&E-only pretraining. Importantly, joint alignment enhances SIM performance without degrading H\&E representations, demonstrating asymmetric enrichment rather than representational averaging. These results suggest that histology-guided cross-modal pretraining yields biologically grounded SIM embeddings suitable for broad downstream reuse.


\end{abstract}

\section{Introduction}
Modern pathology relies on histological staining to reveal cellular and tissue morphology for diagnosis. Conventional slide preparation, however, is slow, destructive, and labor-intensive, limiting its utility for rapid intraoperative or bedside evaluation \cite{Kondepudi2025}. This has motivated the development of label-free or minimal-preparation imaging approaches for fresh tissue—often termed ex vivo microscopy
\cite{dobbs2013feasibility,behr2021structured,behr2024simplosone,snuderl2013dye,stelzer2021light,wang2018partial,fereidouni2017muse}.
Structured Illumination Microscopy (SIM) has emerged as a promising ex vivo modality due to its speed, high contrast, and nondestructive imaging of tissue surfaces \cite{behr2021structured}. By using patterned illumination to achieve wide-field optical sectioning, SIM enables rapid acquisition of large tissue areas and can generate diagnostic-quality images 
within minutes, while preserving tissue for subsequent processing \cite{johnson2020artifact,behr2024simplosone}. These properties position SIM as a practical candidate for real-time, slide-free pathology during surgery.

While SIM enables rapid surface imaging without staining, H\&E remains the diagnostic gold standard due to its rich nuclear and architectural detail and well-established interpretive criteria \cite{Li2025}. By distinctly highlighting nuclei and extracellular structure, H\&E defines a semantically structured morphological space that underpins clinical decision-making. In contrast, SIM relies on fluorescence or reflectance contrast with limited penetration depth ($\sim$50 µm), leading to differences in visual representation and architectural context.
%
%
Rather than treating SIM and H\&E as interchangeable modalities, we leverage H\&E as a semantic prior for representation learning. Prior efforts have attempted to simulate H\&E-like contrast from fresh tissue imaging to preserve diagnostic familiarity \cite{simonson2021creating}. In contrast, we align representations rather than appearances, enabling SIM embeddings to internalize histological structure without requiring stain synthesis. This framing suggests a directional benefit: \textit{SIM representations can be enriched by H\&E-derived priors while preserving modality-specific signal and maintaining stable H\&E performance under joint training.}

At the same time, the rise of foundation models in digital pathology has demonstrated the potential of large-scale self-supervised learning on H\&E data \cite{chen2024towards,lu2024conch,vorontsov2023virchow,karasikov2025training}. However, these models are predominantly trained on conventional brightfield histology and are not explicitly designed to account for thick-tissue fluorescence modalities. Consequently, representations often encode stain-specific color and texture statistics, limiting transfer to substantially different contrast mechanisms such as SIM. For example, H\&E-pretrained models have shown limited gains when transferred to immunohistochemistry domains \cite{Tizhoosh2025,Gallagher-Syed2024}, and recent work on intraoperative label-free microscopy required training large-scale models directly on optical data \cite{Kondepudi2025}.
%
%
%
Existing approaches to stain normalization and domain adaptation typically address modality shift post hoc \cite{Lafarge2019,otalora2019dannpath}, rather than leveraging shared morphological structure during representation learning. We instead treat modality shift as an opportunity for directional cross-modality enrichment, explicitly aligning SIM with H\&E during pretraining to transfer histological priors into SIM embeddings.

\begin{figure}[t]
    \centering
    \includegraphics[width=\linewidth]{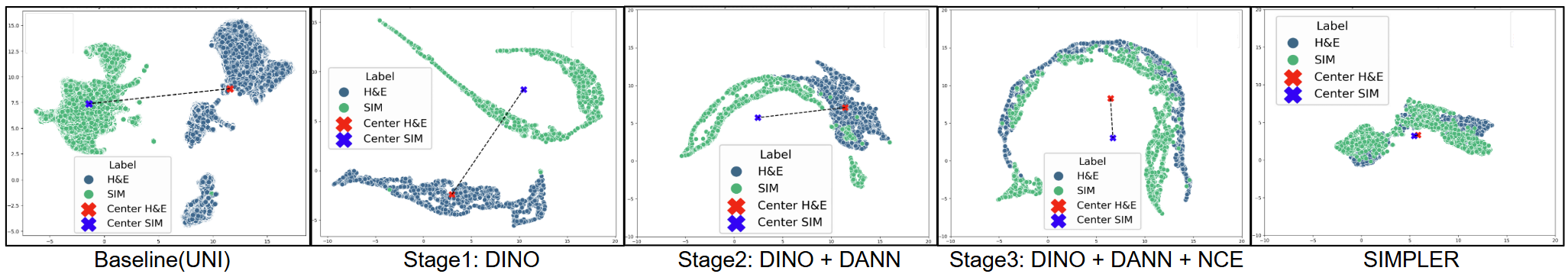}
    \caption{\textbf{UMAP of feature embeddings.} Each stage progressively aligns H\&E and SIM features, reducing modality gap and enhancing shared representation.}
\label{fig:umap}
\vspace{-1.5em}
\end{figure}

To address this gap, we introduce SIMPLER, a self-supervised cross-modality pretraining framework designed to align and enrich representations across H\&E and SIM.
Beginning with self-distillation for robust feature learning, the framework incrementally incorporates domain-adversarial alignment, patch-level contrastive correspondence across paired H\&E–SIM regions, and cross-reconstruction objectives that require embeddings from one modality to predict the other. Together, these stages promote the encoding of histological structure while reducing reliance on superficial modality-specific cues.
Through this pretraining strategy, SIMPLER produces reusable embeddings in which SIM representations are informed by H\&E-derived morphology while maintaining stable H\&E performance. Empirically, we observe directional gains: \textit{improved SIM performance across weakly supervised whole-slide classification and unsupervised morphological clustering, without degradation of H\&E representations}. These findings demonstrate that cross-modality alignment can serve as a principled mechanism for histology-informed enrichment of SIM embeddings.

\vspace{0.05in}
\noindent Our contributions are as follows:
\begin{itemize}
    \item A self-supervised cross-modal framework that aligns SIM and H\&E to learn reusable embeddings spanning fluorescence and brightfield histology.
    \item A progressive curriculum combining distillation, adversarial, contrastive, and reconstruction objectives to reduce modality bias while preserving structure.
    \item Empirical evidence of directional cross-modal enrichment, improving SIM performance without degrading H\&E representations and outperforming state-of-the-art pathology foundation models on weakly supervised classification and unsupervised clustering.
\end{itemize}


\vspace{-1.0em}
\section{Methods}


We address representation transfer between brightfield H\&E and fluorescence SIM, which differ markedly in contrast and statistics. We propose SIMPLER, a progressive self-supervised curriculum that incrementally aligns stained and unstained modalities while preserving modality-specific signal. Built on self-distillation, successive alignment constraints suppress modality cues and promote shared morphology. Figure 2 summarizes the four-stage framework.




\vspace{0.05in}
\noindent\textbf{Stage 1: DINO-Based Self-Distillation.} 
We begin with transformer-based self-supervised pretraining using DINO~\cite{oquab2023dinov2}, a framework widely adopted in pathology foundation models \cite{vorontsov2023virchow,lu2024conch}. It promotes invariance to spatial and photometric perturbations by matching global and local views in a student–teacher setup. We also incorporate batch-style standardization ~\cite{scalbert2023towards} to increase diversity. 



For each image (H\&E or SIM), we generate $M$ global crops and $N$ local crops. 
The teacher $f_{\theta'}$ (momentum encoder) produces temperature-scaled outputs, while the student $f_\theta$ learns to match these distributions via cross-entropy:
{\small
\begin{equation}
   {\mathcal{L}_{\text{DINO}} = \sum_{i \neq j} \text{CE} \left( \text{softmax}\left(\frac{f_{\theta'}(v_i) - c}{T_t}\right), \text{softmax}\left(\frac{f_{\theta}(v_j)}{T_s}\right) \right)}
\end{equation}
}
where $v_i$ and $v_j$ are distinct views of the same patch, $T_t$ and $T_s$ are teacher and student temperatures, and $c$ is a learned centering vector. By enforcing view consistency, DINO encodes coherent morphological structure rather than superficial pixel statistics, but does not address the domain gap.

\begin{figure}[t]
    \centering
    \includegraphics[width=\linewidth]{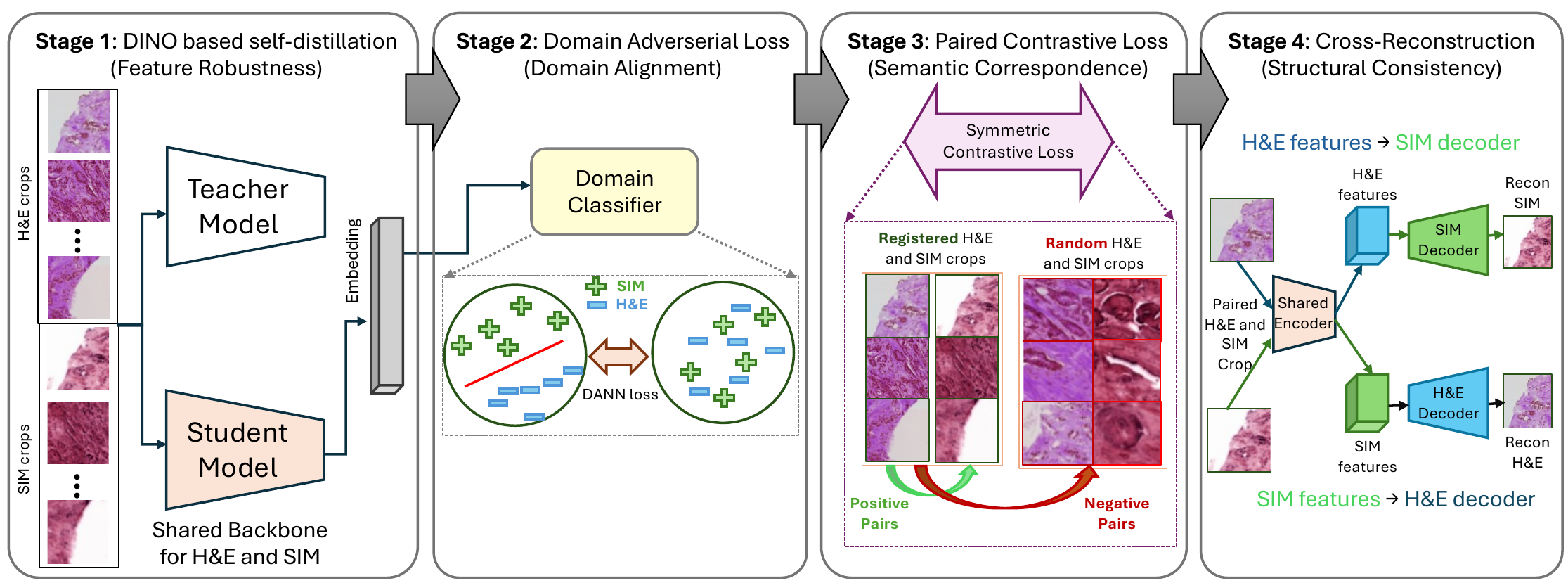}
    \caption{\textbf{SIMPLER Overview.} The proposed progressive cross-modality curriculum used to align H\&E and SIM representations. Each stage introduces increasingly structured alignment constraints, from feature robustness to structural cross-reconstruction.
    }
    \label{fig:methods}
    \vspace{-1.0em}
\end{figure}


 

\vspace{0.05in}
\noindent\textbf{Stage 2: Domain-Adversarial Alignment.}
After Stage 1, embeddings capture rich within-modality morphology but remain separable due to modality-specific imaging physics. Direct cross-modality contrastive alignment at this stage would require pulling semantically corresponding samples across distant feature manifolds, leading to unstable optimization. Stage 2, therefore, introduces coarse distribution-level alignment by connecting a domain classifier $D_\phi$ to the encoder via a gradient reversal layer (GRL)~\cite{ganin2016dann}. The GRL acts as an identity during forward propagation and inverts gradients during backpropagation, enabling adversarial training.
The domain loss is defined as:

{\small
\begin{equation}
    \mathcal{L}_{\text{domain}} = \text{CE}\left(D_\phi(\text{GRL}(f_\theta(v))), y_d\right)
\end{equation}
}
where $v$ is a view from either modality and $y_d$ denotes the domain label. This loss reduces separability between $p(f_\theta(v)\mid y_d = \text{H\&E})$ and $p(f_\theta(v)\mid y_d = \text{SIM})$, providing smoother representation geometry for semantic alignment in Stage 3.






\vspace{0.05in}
\noindent\textbf{Stage 3: Paired Contrastive Alignment for Semantic Consistency.}
While Stage 2 reduces global domain separability, it does not ensure that spatially corresponding regions across H\&E and SIM are aligned. Leveraging spatially registered image pairs, we introduce a patch-wise paired contrastive objective~\cite{chen2020simclr} to enforce instance-level semantic correspondence. Given paired crops from aligned regions, we compute embeddings $z_{he}$ and $z_{sim}$ using $f_\theta$ and minimize. Given paired crops from aligned regions, we compute embeddings 
$z^{i}_{he}$ and $z^{i}_{sim}$ using $f_\theta$ for each paired sample 
$i = 1, \dots, B$ in a minibatch of $B$ registered H\&E-SIM pairs. 
We define the candidate sets: \(\mathcal{Z}^{(i)} = 
\left( \{z^{j}_{he}\}_{j=1}^{B} \cup 
       \{z^{j}_{sim}\}_{j=1}^{B} \right) 
\setminus \{z^{i}_{he}\}\) and \(\tilde{\mathcal{Z}}^{(i)} = 
\left( \{z^{j}_{he}\}_{j=1}^{B} \cup 
       \{z^{j}_{sim}\}_{j=1}^{B} \right) 
\setminus \{z^{i}_{sim}\}\),
where $\mathcal{Z}^{(i)}$ (resp. $\tilde{\mathcal{Z}}^{(i)}$) contains 
all embeddings from the minibatch across both modalities, excluding 
the anchor embedding itself.
The symmetric paired contrastive loss is then:
{\small
\begin{equation}
\mathcal{L}_{\text{contrast}}
=
-\frac{1}{2B} \sum_{i=1}^{B}
\Big[
\ell(z^{i}_{he}, z^{i}_{sim})
+
\ell(z^{i}_{sim}, z^{i}_{he})
\Big]
\end{equation}
}
where $\ell(a,b)=\log\frac{\exp(\mathrm{sim}(a,b)/\tau)}{\sum_{z \in \mathcal{Z}_a}\exp(\mathrm{sim}(a,z)/\tau)}$, $\tau$ is a temperature parameter and $\mathrm{sim}(\cdot,\cdot)$ denotes cosine similarity.
The loss is applied symmetrically across modalities. For each registered H\&E-SIM pair, embeddings form positives, while unrelated crops act as negatives. 
This objective introduces a correspondence-level inductive bias that aligns matched regions across modalities while preserving discrimination among unrelated structures.
Because Stage 2 already reduces global domain separability, contrastive alignment operates in a space where modality manifolds are closer, advancing directional enrichment.

\vspace{0.05in}
\noindent\textbf{Stage 4: Modality-Specific Decoders for Cross-Reconstruction.}
While contrastive alignment enforces correspondence between paired embeddings, it does not ensure that the latent space retains sufficient structural information to explain both modalities. To further disentangle shared morphology from modality-specific appearance, we introduce two modality-specific decoders, $g_{he}$ and $g_{sim}$, that perform cross-reconstruction between H\&E and SIM.


Given a paired patch $(v_{he}, v_{sim})$ and corresponding embeddings $(z_{he}, z_{sim})$ from the shared encoder $f_\theta$, we minimize the cross-reconstruction loss:

\begin{equation}
    \mathcal{L}_{\text{recon}} = \|g_{he}(z_{sim}) - v_{he}\|_2^2 + \|g_{sim}(z_{he}) - v_{sim}\|_2^2
\end{equation}
where $v_{he}$ and $v_{sim}$ denote pixel-level patch intensities. This objective introduces a cross-modal predictive inductive bias: embeddings must encode structural information to sufficiently reconstruct the opposite modality.
Cross-reconstruction regularizes the latent space to preserve dense morphological content while decoders capture modality-specific appearance. Consequently, the encoder is encouraged to retain histological structure that generalizes across imaging physics, reinforcing the directional enrichment of SIM representations.


\noindent\textbf{Progressive Cross-Modality Curriculum.}
Each stage is introduced sequentially, with pretrained weights 
initialized from the previous stage. Specifically, Stage 1 optimizes 
$\mathcal{L}_{\text{DINO}}$ alone. Stage 2 adds the domain-adversarial 
loss, Stage 3 further incorporates the paired contrastive objective, 
and Stage 4 introduces the cross-reconstruction constraint. 
In the final stage, the combined objective is:
\begin{equation}
\mathcal{L}_{\text{total}} 
= \lambda_1 \mathcal{L}_{\text{DINO}} 
+ \lambda_2 \mathcal{L}_{\text{domain}} 
+ \lambda_3 \mathcal{L}_{\text{contrast}} 
+ \lambda_4 \mathcal{L}_{\text{recon}}.
\end{equation}





\vspace{-0.25em}
\noindent\textbf{Implementation Details.}
We adopt a ViT-L/16 backbone~\cite{dosovitskiy2020image} as the shared encoder, with separate projection heads for DINO and contrastive objectives. 
Modality-specific decoders consist of four-layer transposed convolutional blocks. We train on spatially registered H\&E–SIM patch pairs extracted from whole-slide images using TRIDENT [27], sampled at $256 \times 256$ resolution $(20\times)$. Each minibatch consists of paired H\&E and SIM crops to support cross-modality alignment. Optimization uses AdamW (lr $1\times10^{-4}$, batch size 24) across 4 H200 GPUs.
Hyperparameters $\lambda_1$–$\lambda_4$ are tuned stage-wise. Each stage is trained for approximately two days before proceeding to the next.


\section{Results}

\subsection{Dataset}
\textbf{Pretraining cohort.} For cross-modality pretraining, we used 28 spatially registered H\&E–SIM prostate whole-slide image (WSI) pairs. 
From these paired slides, approximately 300{,}000 $256 \times 256$ patches were sampled at $20\times$ magnification after tissue segmentation. Patches used for contrastive and reconstruction objectives were drawn from spatially matched regions.

\noindent\textbf{Evaluation cohort.} To evaluate downstream performance and cross-modal generalization, we use an independent held-out dataset of 115 H\&E and 115 SIM WSIs from 49 patients, with no overlap with pretraining. Slides are not spatially paired or registered across modalities, providing a realistic setting for weakly supervised MIL and unsupervised morphological clustering. 


\newcommand{\cmark}{\checkmark}
\newcommand{\xmark}{\ding{55}}

\begin{table*}[t]
\centering
\begin{minipage}{0.48\textwidth}
\centering
\caption{MIL performance (Accuracy and AUC) on H\&E and SIM domains across different foundation models.}
\label{table:mil_results}
\scalebox{0.80}{
\begin{tabular}{|c|c|c|c|c|c|}
\hline
\multirow{2}{*}{\textbf{Model}} & \multicolumn{2}{c|}{\textbf{H\&E}} & \multicolumn{2}{c|}{\textbf{SIM}} \\ \cline{2-5}
& \textbf{Acc} & \textbf{AUC} & \textbf{Acc} & \textbf{AUC} \\ \hline
UNI \cite{chen2024towards} & \textbf{0.733} & \textbf{0.722} & 0.665 & 0.700 \\
Conch \cite{lu2024conch} & 0.550 & 0.600 & 0.466 & 0.550 \\
Midnight12K \cite{karasikov2025training} & 0.600 & 0.655 & 0.533 & 0.600 \\
SIMPLER & \textbf{0.733} & 0.704 & \textbf{0.867} & \textbf{0.833} \\ \hline
\end{tabular}}
\end{minipage}
\hfill
\begin{minipage}{0.50\textwidth}
\centering
\caption{Ablation study on SIM domain showing the effect of progressively adding each component of SIMPLER.}
\label{table:ablation}
\scalebox{0.77}{
\begin{tabular}{|c|c|c|c|c|c|c|c|}
\hline
\textbf{SIM} & \textbf{H\&E} & \textbf{DINO} & \textbf{DANN} & \textbf{NCE} & \textbf{Recon} & \textbf{Acc} & \textbf{AUC} \\ \hline
\cmark & \xmark & \cmark & \xmark & \xmark & \xmark & 0.766 & 0.736 \\ \hline
\cmark & \cmark & \cmark & \xmark & \xmark & \xmark & 0.665 & 0.667 \\
\cmark & \cmark & \cmark & \cmark & \xmark & \xmark & 0.700 & 0.667 \\
\cmark & \cmark & \cmark & \cmark & \cmark & \xmark & 0.800 & 0.784 \\
\cmark & \cmark & \cmark & \cmark & \cmark & \cmark & \textbf{0.867} & \textbf{0.833} \\ \hline
\end{tabular}
}
\end{minipage}
\vspace{-0.5em}
\end{table*}


\vspace{-1em}
\subsection{Downstream Task: Multiple Instance Learning (MIL)}

We evaluate the quality of learned representations using a binary whole-slide classification task under an attention-based MIL framework~\cite{ilse2018abmil}. The evaluation cohort was split into 85 WSIs for training, 15 for validation, and 15 for testing. Separate MIL classifiers were trained for H\&E and SIM modalities using frozen features extracted from each pretrained backbone.

\noindent\textbf{Feature Space Alignment.}
Figure~\ref{fig:umap} visualizes UMAP projections of patch-level embeddings. Features from UNI \cite{chen2024towards} exhibit clear modality-dependent clustering, with H\&E and SIM occupying distinct regions. In contrast, successive stages of SIMPLER progressively reduce this separation, resulting in increased overlap between modalities while preserving intra-class structure. This progression reflects the intended curriculum effect: marginal alignment (Stage 2) followed by instance-level semantic anchoring (Stage 3).


\noindent\textbf{Quantitative Performance.}
Table~\ref{table:mil_results} reports MIL accuracy and AUC for both domains. SIMPLER achieves substantial improvements on the SIM domain (Acc = 0.867, AUC = 0.833), outperforming all compared foundation models. Importantly, H\&E performance remains competitive with UNI (Acc = 0.733 vs. 0.733), indicating that cross-modality alignment enhances SIM generalization without degrading performance on the reference modality. This asymmetric improvement supports our directional enrichment hypothesis: histological priors inform SIM representations while preserving H\&E fidelity.

\begin{figure}[t]
    \centering
    \includegraphics[scale=0.20]{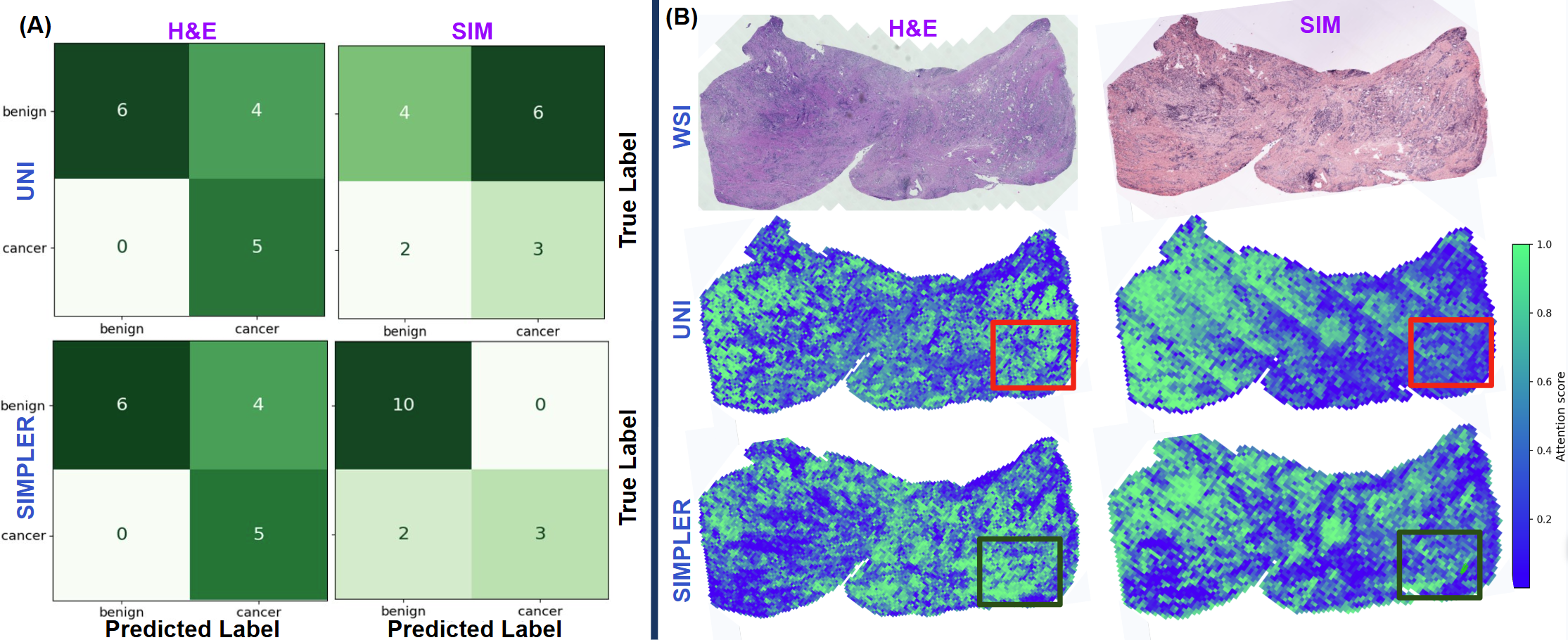}
    \caption{\textbf{Results.} (A) Confusion matrices for H\&E and SIM domains comparing UNI and SIMPLER.  (B) Qualitative heatmap comparison of attention scores. }
    \label{fig:mil_results}
    \vspace{-1em}
\end{figure}


\noindent\textbf{Qualitative Analysis.}
Figure~\ref{fig:mil_results} provides qualitative comparisons. Panel (A) shows confusion matrices for H\&E and SIM domains, highlighting reduced misclassification in SIM for SIMPLER relative to UNI. Panel (B) compares attention heatmaps. For visualization purposes only, H\&E and SIM slides were manually aligned to better illustrate spatial correspondence. SIMPLER produces more spatially coherent and morphologically plausible attention regions (green boxes) compared to UNI (red boxes), suggesting improved localization of diagnostically relevant structures.


\noindent \textbf{Ablation Studies.}
Table~\ref{table:ablation} evaluates each stage of the SIMPLER curriculum. Starting from a DINO baseline, progressively adding domain-adversarial alignment, paired contrastive correspondence, and cross-reconstruction yields monotonic SIM gains. The largest boost comes from paired contrastive alignment, with cross-reconstruction providing further refinement. A SIM-only model with DINO objective fails to match SIMPLER’s performance. These results validate the hierarchical inductive bias: \textit{global domain alignment enables stable semantic correspondence, while reconstruction regularizes the embedding space for cross-modal predictiveness.}

\begin{figure}[t]
    \centering
    \includegraphics[scale=0.20]{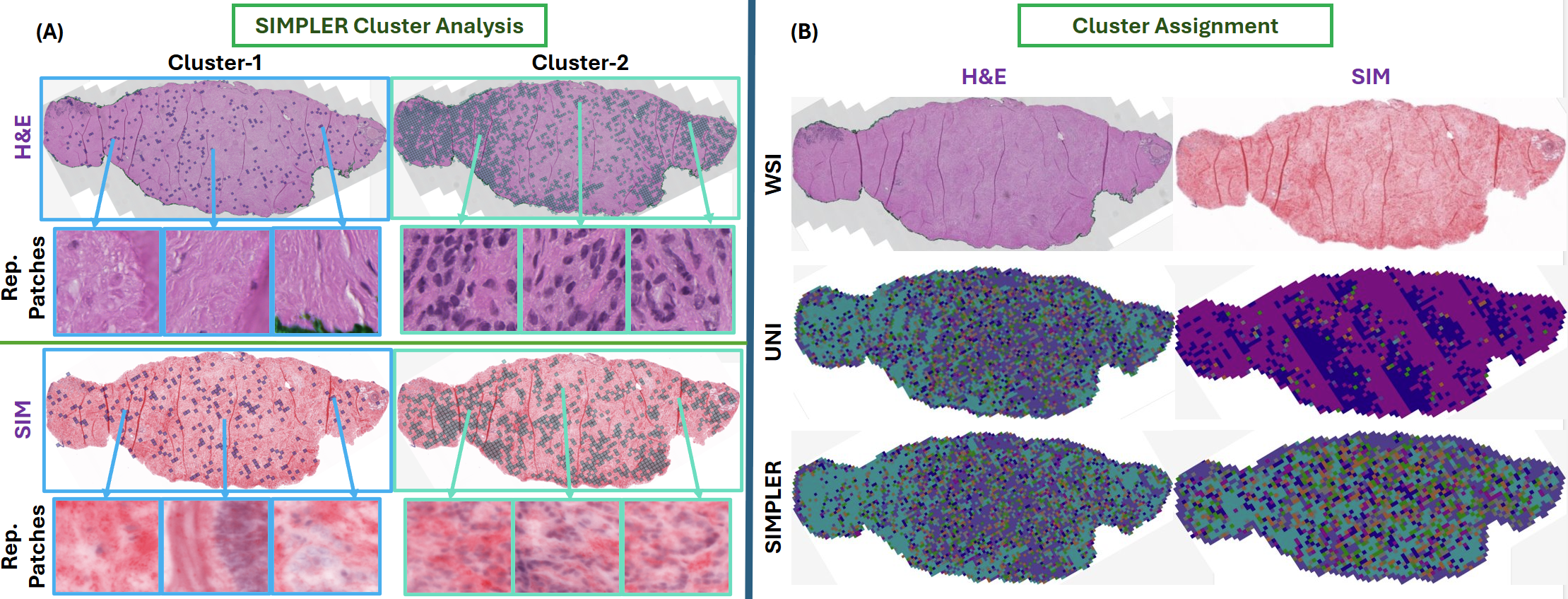}
    \caption{\textbf{Clustering results.}(A) Representative patches from two clusters identified by SIMPLER show consistent structure. (B) Cluster maps using UNI (middle) and SIMPLER (bottom). SIMPLER achieves strong cross-modal consistency}
    \label{fig:clustering_results}
    \vspace{-1em}
\end{figure}

\vspace{-1.0em}
\subsection{Downstream Task: Morphological Clustering}

To evaluate whether learned representations preserve morphologically meaningful structure across modalities, we perform unsupervised morphological clustering following the PANTHER framework~\cite{ren2024panther}. This approach groups patch-level embeddings into discrete morphological prototypes (e.g., tumor nests, glandular structures, stroma), enabling analysis of dominant tissue phenotypes without supervision.
%
We extract patch-level embeddings from the same held-out evaluation cohort used for MIL. Importantly, this dataset contains no spatial registration between H\&E and SIM slides. For each modality independently, we fit a Gaussian Mixture Model (GMM) to partition the feature space into $N$ (N=8) clusters. Because clustering is performed separately per modality and without cross-modal correspondence, any similarity in cluster structure across domains arises solely from alignment in the learned representation space.



Figure~\ref{fig:clustering_results}(B) compares cluster assignments across H\&E and SIM whole-slide images. UNI shows modality-specific cluster patterns, indicating divergent embedding geometries. In contrast, SIMPLER produces consistent spatial cluster distributions across modalities, suggesting better preservation of shared morphology. To interpret cluster semantics, we select representative patches via 
$k$-medoids within each cluster. Figure~\ref{fig:clustering_results}(A) presents examples from two dominant clusters: Cluster-2 captures densely packed nuclear regions in both modalities, whereas Cluster-1 corresponds to lower cellularity or stromal areas.


%
%
These findings indicate that SIMPLER enables consistent unsupervised grouping of tissue phenotypes across stained and unstained domains, reinforcing the directional enrichment effect observed in downstream classification.

\vspace{-1.0em}
\section{Conclusion}
\vspace{-0.5em}
We introduced SIMPLER, a cross-modality self-supervised framework that aligns surface-based imaging, such as SIM, with traditional histopathology (H\&E) through histology-informed representation learning. 
Using a progressive four-stage curriculum self-distillation, domain-adversarial alignment, paired contrastive correspondence, and cross-reconstruction, we achieve consistent SIM gains while preserving H\&E fidelity. Across downstream tasks, including MIL classification and unsupervised clustering, SIMPLER demonstrates directional enrichment: SIM benefits from histological priors without degrading the reference modality.
These findings suggest that structured cross-modal alignment, rather than modality-specific fine-tuning or stain simulation, offers a principled strategy for learning biologically grounded representations across diverse histological conditions. This reframes modality shift as an opportunity for structured representation transfer rather than a constraint of stain-specific modeling.



\section{Acknowledgements}
\sloppy{Research reported in this publication was supported by the Advanced Research Projects Agency for Health (ARPA-H) under award number D24AC00338-00. The content is solely the responsibility of the authors and does not necessarily represent the official views of the Advanced Research Projects Agency for Health.}
\newpage
\bibliographystyle{splncs04}
\bibliography{mybibliography}




\end{document}